\documentclass[letterpaper, 10 pt, conference]{ieeeconf}

\IEEEoverridecommandlockouts

\usepackage{color}
\usepackage{xcolor}
\usepackage{soul}
\usepackage{multirow}
\usepackage{booktabs}
\usepackage{url}
\usepackage{subfigure}
\usepackage{threeparttable}
\usepackage{cite}
\makeatletter
\let\NAT@parse\undefined
\makeatother
\usepackage{hyperref}[colorlinks,linkcolor=blue]
\usepackage[ruled,linesnumbered]{algorithm2e}
\renewcommand{\baselinestretch}{0.966}
\newcommand{\cmmnt}[1]{\ignorespaces}

\usepackage{booktabs}
\usepackage[margin=1in]{geometry}
\usepackage{array}
\newcolumntype{L}[1]{>{\raggedright\arraybackslash}p{#1}}

\usepackage{enumerate}
\usepackage{amssymb}
\usepackage{leftidx}
\usepackage{amsfonts}
\usepackage{amsmath}
\usepackage{bm}
\usepackage{booktabs}
\usepackage{setspace}
\usepackage{makecell}
\usepackage{setspace}
\usepackage{float}
\usepackage[pdftex]{graphicx}
\usepackage{cite}
\usepackage{siunitx}

\graphicspath{{figs/}}
\usepackage{graphicx}
\graphicspath{{figs/}}

\makeatletter
\newcommand*{\rom}[1]{\expandafter\@slowromancap\romannumeral #1@}
\makeatother
\usepackage[export]{adjustbox}

\begin{document}

\title{\LARGE \bf
Bubble Explorer: Fast UAV Exploration in Large-Scale and Cluttered 3D-Environments using Occlusion-Free Spheres
}

\author{Benxu~Tang*, Yunfan~Ren*, Fangcheng~Zhu, Rui~He, Siqi~Liang, Fanze~Kong and Fu~Zhang
 \thanks{*These two authors contributed equally to this work.}
 \thanks{B. Tang, Y. Ren, F. Zhu, R. He, F. Kong and F. Zhang are with the Department of Mechanical Engineering, University of Hong Kong
  \texttt{\{tangbenx, renyf,  zhufc, herui, kongfz\}}\texttt{@connect.hku.hk},
  \texttt{fuzhang}\texttt{@hku.hk}, S. Liang is with School of Mechanical Engineering and Automation, Harbin Institute of Technology 
 \texttt{sqliang@stu.hit.edu.cn}.
}
}

\maketitle
\pagestyle{empty} 
\thispagestyle{empty} 

\begin{abstract}
Autonomous exploration is a crucial aspect of robotics that has numerous applications. Most of the existing methods greedily choose goals that maximize immediate reward. This strategy is computationally efficient but insufficient for overall exploration efficiency. 
In recent years, some state-of-the-art methods are proposed, which generate a global coverage path and significantly improve overall exploration efficiency. However, global optimization produces high computational overhead, leading to low-frequency planner updates and inconsistent planning motion. In this work, we propose a novel method to support fast UAV exploration in large-scale and cluttered 3-D environments. We introduce a computationally low-cost viewpoints generation method using novel occlusion-free spheres. Additionally, we combine greedy strategy with global optimization, which considers both computational and exploration efficiency.
We benchmark our method against state-of-the-art methods to showcase its superiority in terms of exploration efficiency and computational time. We conduct various real-world experiments to demonstrate the excellent performance of our method in large-scale and cluttered environments.

\end{abstract}

\section{Introduction}
\label{sec:intro}
Autonomous exploration, where robots explore unknown environments and gather information independently, has become increasingly popular in applications such as mine exploration, industrial inspection, and search and rescue operations.
Robots can access areas that are difficult for humans to reach, and reduce the risks humans expose to in hazardous environments.

The task of autonomous exploration is to plan a path to explore the entire unknown environment as quickly as possible. Various exploration methods have been proposed in recent years to tackle the task. Most of these methods adopt a greedy strategy. ~\cite{bircher2016receding, dang2019graph, dharmadhikari2020motion} span RRT~\cite{lavalle1998rapidly} in the environment and select the node with the highest information gain to visit.~\cite{yamauchi1997frontier, cieslewski2017rapid} select the frontier that minimizes the traversal cost or the direction change of the UAV as the goal. The greedy-based methods are computationally efficient but insufficient in terms of overall exploration efficiency, as they ignore global optimality and generate back-and-forth movements. Other methods, such as~\cite{zhou2021fuel}, adopt a global optimization strategy that finds a global tour to visit unexplored regions. This strategy improves overall exploration efficiency but results in high computational time, leading to low planner update frequency and inconsistent planning motion.
Moreover, existing methods generate viewpoints in a sampling way and evaluate the reward of the viewpoint using a computationally expensive ray-casting process, which further increases the computational cost. 

\begin{figure}[t]
 \centering 
\includegraphics[width=0.47\textwidth]{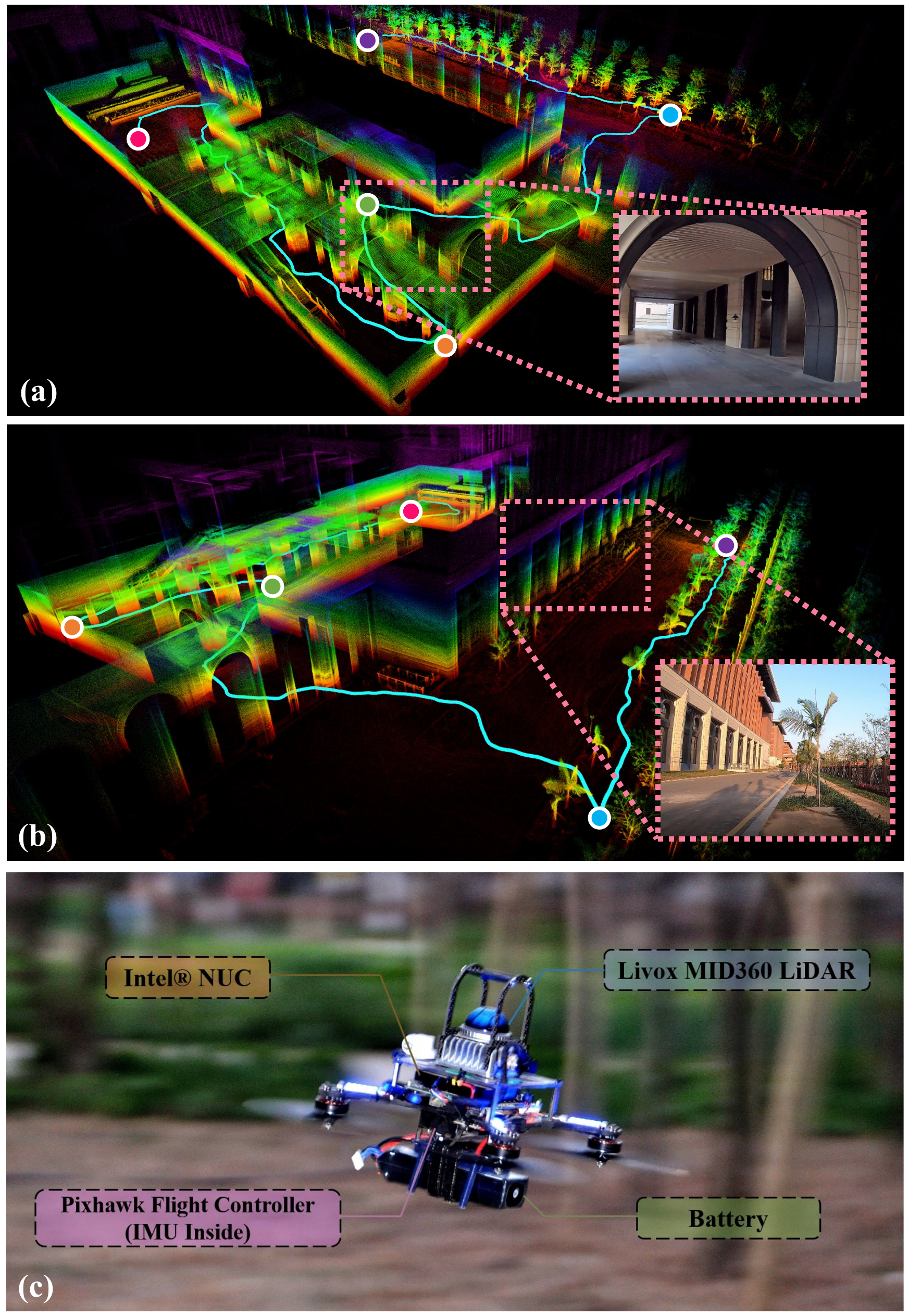}
 \caption{Performing exploration task in a large-scale environment composed of both indoor and outdoor spaces. (a) and (b): Two different views of the online-built point cloud map, and trajectory executed by the UAV (light blue line), with images displaying the environment. The points with the same color indicate the same position. (c): The quadrotor platform used in the exploration.}
 \label{fig:cover}
\end{figure}

Motivated by these facts, we propose a novel method that can support fast and efficient UAV exploration in large-scale and cluttered 3-D environments. We introduce two key contributions: 1) A novel concept of the occlusion-free sphere, which generates high-quality viewpoints at a low computational cost. 2) Based on the generated viewpoints, we introduce a novel strategy that combines greedy with global optimization, which finds an efficient global tour visiting high-gain viewpoints, balancing overall exploration efficiency and computational cost. Finally, we design a local planner that generates safe and kinodynamically feasible trajectories for the UAV to follow. We validate the proposed method through extensive simulation and real-world experiments, showing that it outperforms the state-of-the-art baselines in terms of both exploration efficiency and computational time.

To sum up, the contributions of this paper are listed below:

\begin{itemize}
 \item [1)] 
 We propose a novel concept of the occlusion-free sphere to generate high-quality viewpoints, which significantly saves computational time and improves exploration efficiency.
 
 \item [2)] 
 Based on the generated viewpoints, we introduce a novel strategy that combines greedy and global optimization, which finds an efficient global tour to visit high-gain unexplored regions, balancing overall exploration efficiency and computational cost. 
 
 \item [3)]
 Extensive simulation experiments demonstrate the advantages of the proposed planner over the state-of-the-art baselines, in terms of exploration efficiency and computational time. 

 \item [4)]
 Implementation of the proposed planner on a fully autonomous quadrotor platform. Various real-world tests show the outstanding performance of the proposed planner in large-scale and cluttered real-world environments.

\end{itemize}

\section{Related Works}
\label{sec:related}

Autonomous exploration has been an active area of research in recent years, and a variety of methods have been proposed to tackle the problem. Sampling-based exploration~\cite{bircher2016receding, dang2019graph, dharmadhikari2020motion, schmid2020efficient} is one of the classic approaches. The approach spans a Rapidly-exploring Random Tree (RRT)~\cite{lavalle1998rapidly} in free space. It evaluates the information gain of the nodes in RRT by the coverage of the unknown region, weighted with the traversal cost to reach it from the current position. The coverage is counted by the number of unknown voxels that fall in the sensor field of view (FoV) and are not occluded by occupied voxels (e.g., by ray-casting). The node with the highest gain is selected as the goal and a traversable path to the node is derived from the RRT. This scheme is first introduced by the Next-Best-View Planner (NBVP)~\cite{bircher2016receding}, and further improved by GBP~\cite{dang2019graph} and MBP~\cite{dharmadhikari2020motion}. In GBP~\cite{dang2019graph}, a topological global map is constructed during the local exploration process. When the local area is fully explored, or the vehicle encounters a dead end, the method finds a path on the global map and redirects the vehicle to unexplored areas. MBP~\cite{dharmadhikari2020motion} constructs RRT using motion-primitives and produces smooth trajectories for the vehicle to execute.

Another classic approach is frontier-based exploration~\cite{yamauchi1997frontier, cieslewski2017rapid, julia2012comparison, shen2012autonomous, zhou2021fuel, cao2021tare}. In frontier-based exploration, the vehicle navigates close to the frontier, defined as the boundary between the free and unknown space, to continue exploring the unknown space. This method is first introduced by~\cite{yamauchi1997frontier}, in which the closest frontier is selected as the next goal. To achieve high-speed flight, ~\cite{cieslewski2017rapid} selects the frontier in sensor FoV and minimizes the velocity change of the vehicle.~\cite{selin2019efficient} analyzes the strengths and weaknesses of the sampling-based and frontier-based approaches. It combines them together by improving NBVP~\cite{bircher2016receding} for local exploration and using a frontier-based approach for global exploration.

The above methods are greedy-based, which select goals that maximize the immediate reward to visit at each planning iteration. This strategy is computationally efficient but insufficient in terms of overall exploration efficiency, as it produces back-and-forth planning motions. Fast UAV Exploration planner (FUEL)~\cite{zhou2021fuel} considers the global optimality. It begins by clustering the frontier cells using a region-growing algorithm and performing Principal Component Analysis (PCA) to split large frontier clusters into smaller ones along the first principal axis. Viewpoints are then sampled around the frontier clusters within a cylindrical coordinate system and evaluated by frontier coverage using ray-casting. The viewpoints with the highest frontier coverage are selected. After that, the method finds a global tour that minimizes the global traversal cost, starting from the current vehicle position and passing through all selected viewpoints. It formulates the problem as a variant of the Traveling Salesman Problem (TSP). To solve the problem, the algorithm first searches for collision-free paths between each pair of viewpoints and between each viewpoint and the current vehicle position using the A* algorithm on the voxel grid map. Then, the algorithm evaluates the connection cost based on the length of the collision-free path and composes a $n_v \times n_v$ cost matrix $\mathbb M_{tsp}$. Finally, the problem can be solved using available TSP algorithms~\cite{lin1973effective}. This method outperforms the greedy-based methods in terms of overall exploration efficiency, but performing global optimization in the entire environment incurs high computational overhead, especially in large-scale environments.

In the proposed method, we improve the scheme of FUEL~\cite{zhou2021fuel} further by generating high-quality viewpoints using occlusion-free spheres, and combining greedy and global optimization strategies. We benchmark our method against the state-of-the-art baselines: FUEL~\cite{zhou2021fuel}, GBP~\cite{dang2019graph} and NBVP~\cite{bircher2016receding}

\begin{figure}[t]
 \centering 
\includegraphics[width=0.45\textwidth]{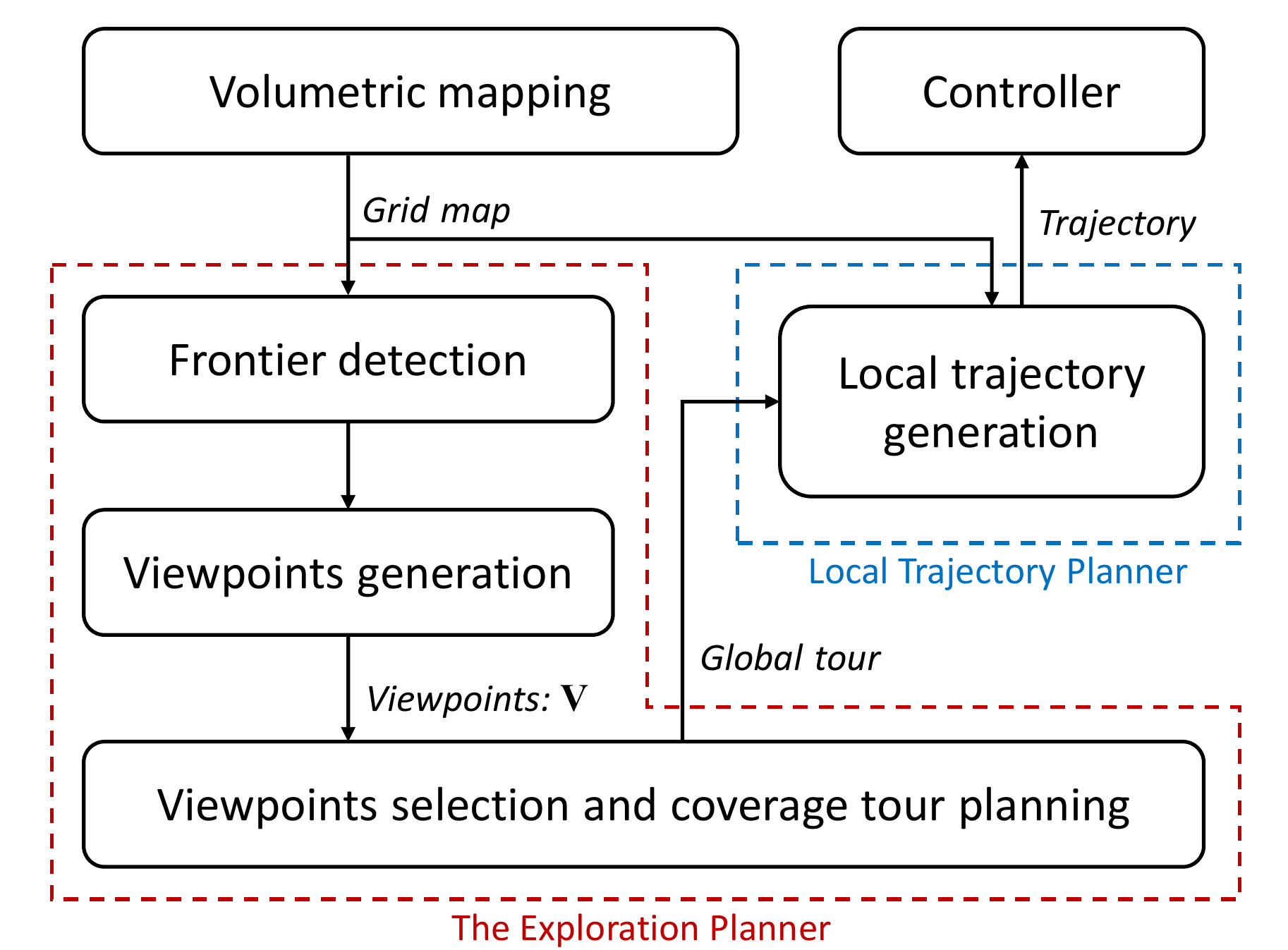}
 \caption{Overview of the proposed system framework}
 \label{fig:overview}
\end{figure}

\section{Proposed Planner}

The overview of the proposed system framework is shown in Fig.~\ref{fig:overview}, including: \textbf{1)} Frontier detection and viewpoints generation using occlusion-free sphere (Sec.~\ref{sec:sphere} and Sec.~\ref{sec:viewpoint}); \textbf{2)} Global exploration tour planning (Sec. \ref{sec:global}); \textbf{3)} Local trajectory generation. (Sec.~\ref{sec:local});

\subsection{Occlusion-Free Sphere}\label{sec:sphere}

\begin{figure}[htbp]
 \centering 
 \includegraphics[width=0.45\textwidth]{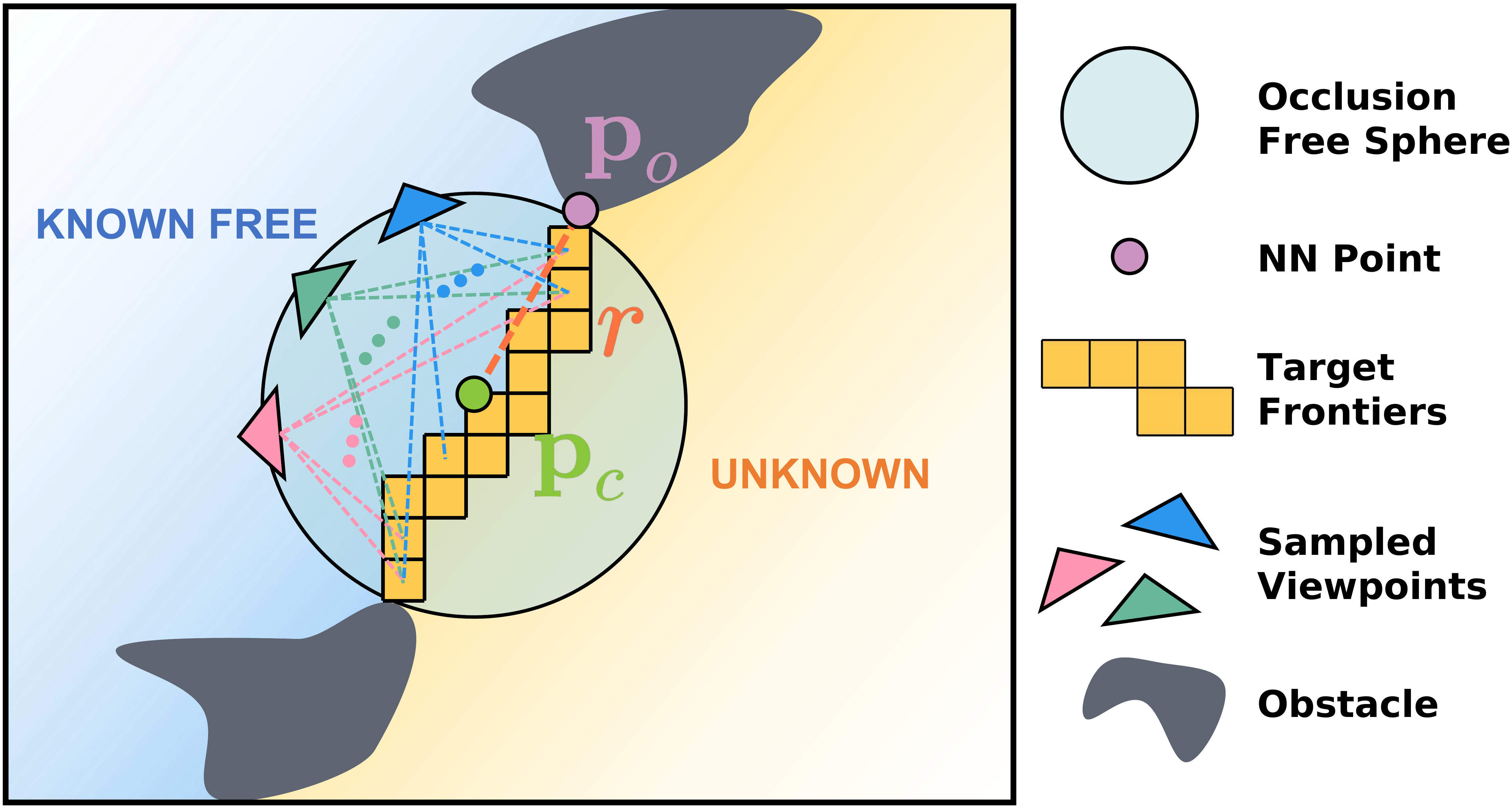}
 \caption{The definition of the occlusion-free sphere.}
 \label{fig:sphere}
\end{figure}

An occlusion-free sphere is defined by its center $ \mathbf p_c \in \mathbb R^3 $, which lies on the target frontier, and the radius:
\begin{equation}
\label{eqa:radius_def}
r = ||\mathbf p_c - \mathbf p_{o}||_{2}
\end{equation}
where $ \mathbf p_{o} \in \mathbb R^3 $ is the nearest neighbor obstacle point (NN point). In this way, the interior of the sphere is free from occupied grids. Since sphere is convex, any line segment that connects points within the sphere (including its surface) and the frontier is occlusion-free, as shown in Fig.~\ref{fig:sphere}. By employing a viewpoint sampling strategy on the sphere's surface, we can obtain high-quality viewpoints without resorting to computationally expensive ray casting techniques.

To expedite the sphere generation process, we adopt an incremental KD-tree methodology as outlined in \cite{cai2021ikd, xu2022fast}. The time complexity of nearest neighbor search (NNS) is $O(\log n)$, where $n$ is the number of nodes in the tree. Therefore, even in large-scale environments, the sphere generation process remains computationally efficient. We denote this process as $\mathtt{GenerateNewSphere}(\mathbf p_c)$, which we shall utilize in subsequent analyses.

\begin{algorithm}[t]
 \small
 \caption{Generate Viewpoints}
 \label{alg:genvps}
 \textbf{Notation}: Input frontier cells $\mathbf F$; Viewpoints $\mathbf V$; Occlusion-free sphere priority queue sort by radius $\mathbf S$; Sphere center list $\mathbf C$; The generated viewpoint $v_b$; The frontier cells covered by $v_b$: $\mathbf F_v$; The sphere centers covered by $v_b$: $\mathbf S_v$\\
 \KwIn{
  $\mathbf F $
 }
 \KwOut{
  $\mathbf V$
 } 
 \BlankLine
 $\mathbf C = \mathtt{DownsampleFrontier}(\mathbf F)$\;\label{alg:genvps:down}
 \For{$\mathbf p_c \in \mathbf C$}{\label{alg:gen_sphere}
 $s_i = \mathtt{ GenerateNewSphere}(\mathbf p_c)$ \;
 $\mathbf S.\mathtt{PushBack}(s_i)$\; 
 }\label{alg:gen_sphere}
 \While{\textbf{not} $\mathbf S.\mathtt{empty}$}{\label{alg:gen_vp}
 $s_l = \mathbf S.\mathtt{front}()$\;
 $\mathbf S.\mathtt{pop}()$\;
 $v_b, \mathbf F_v, \mathbf S_v = \mathtt{GenerateViewpoint}(s_l, \mathbf F $)\;
 $\mathbf F.\mathtt{remove}(\mathbf F_v)$\;
 $\mathbf S.\mathtt{remove}(\mathbf S_v)$;

 $\mathbf V.\mathtt{PushBack}(v_b)$\;\label{alg:gen_vp}
 
 }
\Return $\mathbf V$
\end{algorithm}

\begin{figure}[t]
 \centering 
 \includegraphics[width=0.35\textwidth]{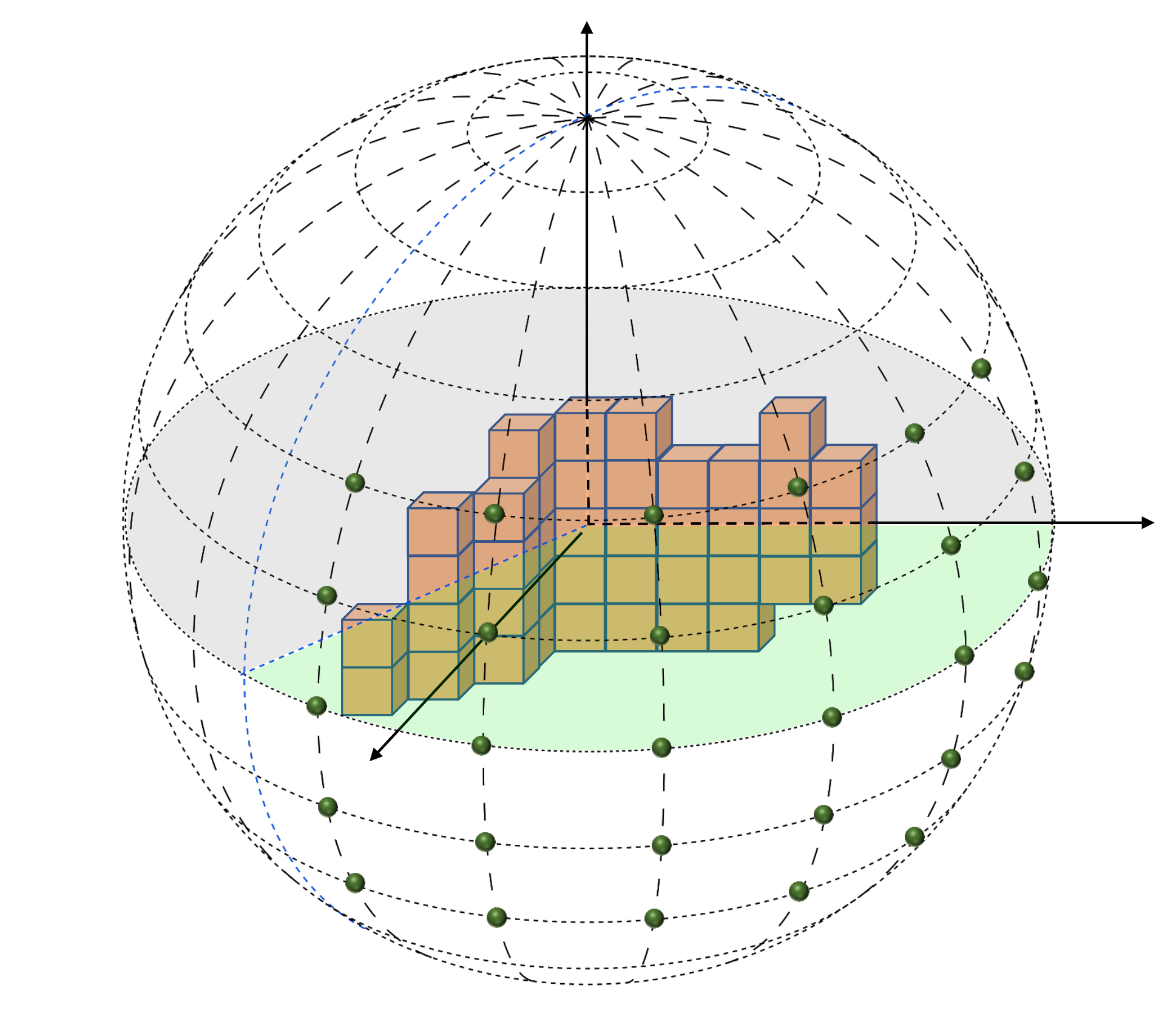}
 \caption{Uniformly sample viewpoints on the sphere surface using a spherical coordinate system. The green points are viewpoint candidates in free space. The orange boxes are frontier cells in sphere.}
 \label{fig:sample}
\end{figure}

\begin{figure*}[h]
 \centering 
 \includegraphics[width=0.97\textwidth]{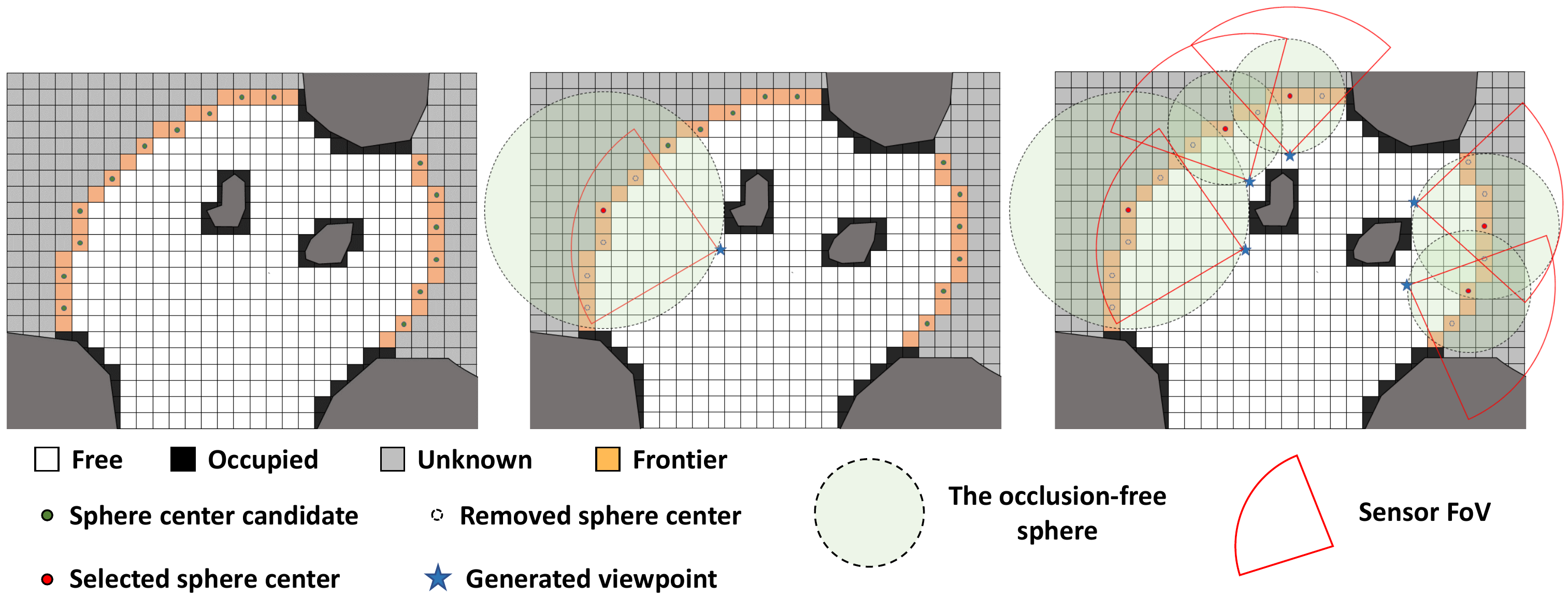}
 \caption{Illustration of viewpoints generation. Left: Uniformly downsample to generate sphere center candidates. Middle: Find the largest sphere and generate a viewpoint. Remove the sphere centers covered by the viewpoint. Right: Perform the same sequence for remaining sphere center candidates and generate a set of viewpoints}
 \label{fig:seqs}
\end{figure*}

\subsection{Viewpoints Generation}
\label{sec:viewpoint}
After generating an occlusion-free sphere $s_l$, the proposed method uniformly samples a set of viewpoints on the sphere surface using a spherical coordinate system, as shown in Fig.~\ref{fig:sample}. Compared to the viewpoint in the sphere, the viewpoint on the sphere surface has a longer viewing distance of the frontier cells inside the sphere, providing more coverage. The yaw direction of the sampled viewpoints is optimized to have the maximum coverage of frontier cells, similar to~\cite{witting2018history}. We then remove the viewpoints in unknown space and perform a sensor FoV check to count the number of frontier cells covered by each remaining viewpoint. Finally, we select the viewpoint that has the highest coverage. This entire process is referred to as $\mathtt{GenerateViewpoint}(s_l, \mathbf F)$, which will also be utilized in subsequent analyses.

Alg.~\ref{alg:genvps} present the workflow of the entire viewpoints generation process. Initially, the algorithm searches for frontier cells in the environment using an incremental manner, similar to FUEL. Then, it uniformly downsamples frontier cells to generate a set of sphere center candidates $\mathbf C$. This process is referred to as $\mathtt{DownsampleFrontier}(\mathbf F)$ (Line~\ref{alg:genvps:down}). Next, for each sphere center $\mathbf p_c$ in $\mathbf C$, the corresponding occlusion-free sphere $s_i$ is generated using $\mathtt{GenerateNewSphere}(\mathbf p_c)$, as described in section \rom{3}.A. The generated sphere $s_i$ is then added to the priority queue $\mathbf S$, which is sorted by sphere radius. 
At each iteration, the largest sphere $s_l$ is selected from the priority queue $\mathbf S$. The algorithm then generates the viewpoint having the highest coverage $v_b$ using $\mathtt{GenerateViewpoint}(s_l, \mathbf F)$, as described in Section~\rom{3}.B. The generated viewpoint $v_b$ covers a set of frontier cells $\mathbf F_v$ as well as a set of sphere centers $\mathbf S_v$. These sets are then removed from $\mathbf F$ and $\mathbf S$, respectively.
The iteration terminates when the sphere list $\mathbf S$ is empty.
Note that if the selected sphere $s_l$ is smaller than a certain threshold, the proposed method generates the viewpoint $v_b$ using a similar approach to FUEL. Specifically, it uniformly samples viewpoints within a spherical coordinate system, with the minimum sampling radius set to be the same as $s_l$ and the maximum radius set to be three times the minimum. In this case, ray-casting is employed to evaluate the frontier coverage of the viewpoints. Note that as only a few frontier cells are contained in the small $s_l$, and the sampling radius is small, the ray-casting process is not computationally intensive.
We refer to the above section as the front-end of the algorithm. Fig.~\ref{fig:seqs} provides an illustration of the entire process.

As described in~\ref{sec:sphere}, the viewpoint generated by our method has occlusion-free coverage of any frontier cell within the sphere. Therefore, the computationally expensive ray-casting process, which is commonly used in existing methods to evaluate sensor coverage, is no longer needed in our approach. This significantly reduces the computational complexity.

A viewpoint located far away from the frontier can cover more frontier cells but is more likely to be occluded by obstacles. In our method, we generate large occlusion-free spheres in open areas, with viewpoints having longer viewing distances to the frontier, covering more frontier cells. In contrast, small spheres and close viewpoints are generated in cluttered spaces to avoid occlusion. Note that both occlusion and short viewing distance reduce the viewpoint coverage.
Exiting methods 
like FUEL require dense sampling in the radius dimension of the cylindrical coordinate system to generate high-coverage viewpoints, resulting in a large number of samples. Note that every sampled viewpoint requires performing ray-casting to evaluate its coverage. This significantly increases the computational time. On the other hand, low-density sampling results in the failure of high-coverage viewpoints generation, leading to reduced exploration efficiency.

\subsection{Global Tour Planning}
\label{sec:global}

We define the gain of a viewpoint $v$ as
\begin{equation}
\label{eqa:gain}
g(v) = r(s)e^{- \lambda c(v, \xi)}
\end{equation}
where $r(s)$ is the radius of the corresponding occlusion-free sphere. $c(v, \xi)$ is the cost going to the viewpoint $v$ from the vehicle current configuration $\xi$. The cost is evaluated using Euclidean distance between $v$ and $\xi$. $\lambda$ is the tuning factor.

A large occlusion-free sphere around the frontier indicates a relatively large volume can be covered without occlusions, hence increasing the chance of discovering more unknown regions. 
A low value of $\lambda$ prioritizes visits to these viewpoints. A high value of $\lambda$ strongly penalizes the distance cost going to the viewpoint. In this case, the planner tends to select close viewpoints and prioritize their visits.

\begin{figure}[t]
 \centering 
 \includegraphics[width=0.45\textwidth]{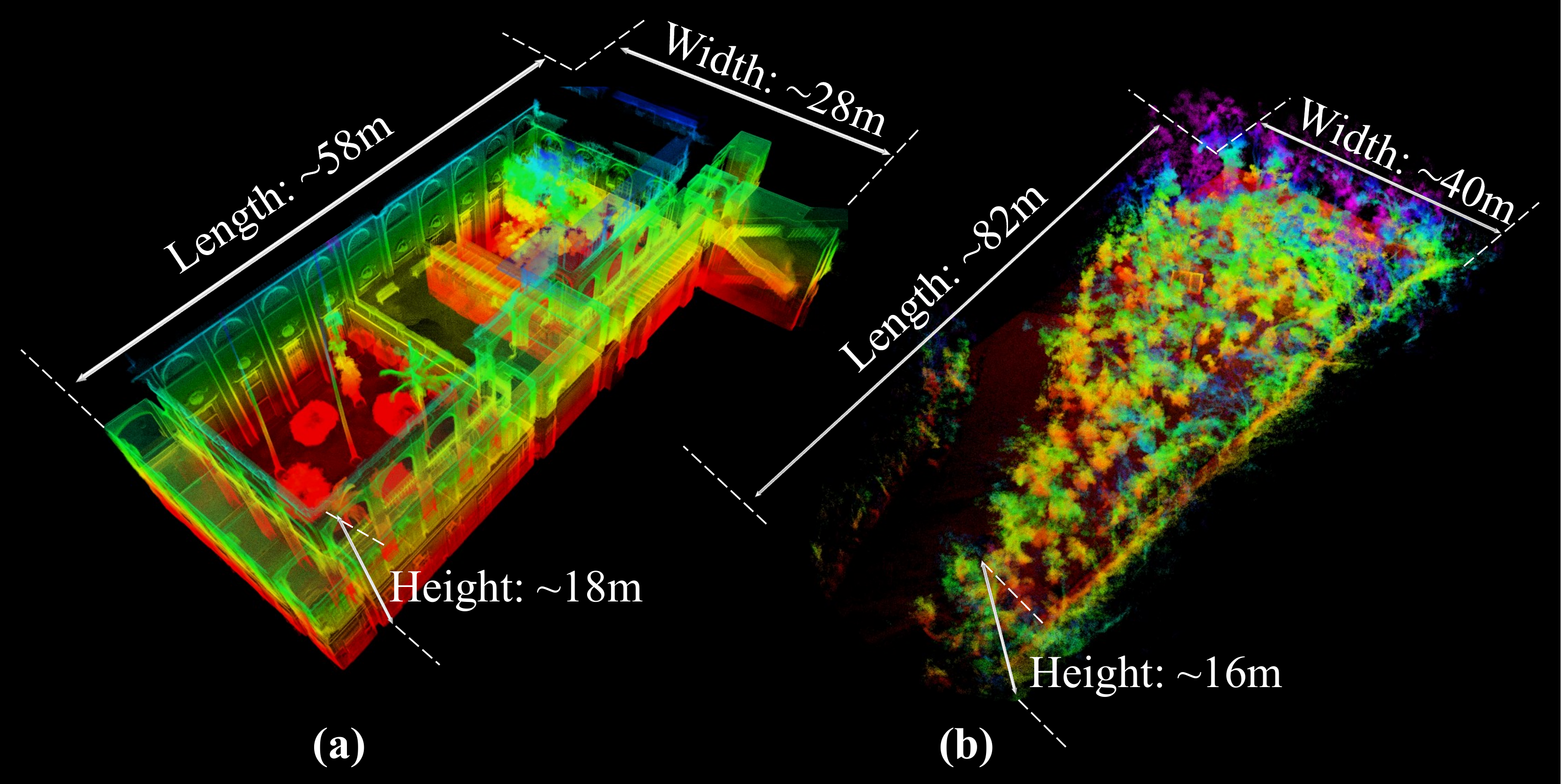}
 \caption{The visualization of the two simulation scenarios. (a): The \textit{Building} scenario. (b): The \textit{Forest} scenario.}
 \label{fig:map}
\end{figure}

The proposed method maintains a viewpoint priority queue $\mathbf Q$, with a fixed size of $n_q$.
As described in section \rom{3}.B, the proposed method generates a set of viewpoints $\mathbf V$ with a total number of $n_v$. For each viewpoint $v_i$ in $\mathbf V$, we compute the gain of $v_i$ defined by~\ref{eqa:gain}. Then we greedily select the viewpoint with the highest gain and push it into priority queue $\mathbf Q$ until the queue is full.

The global planning problem is to find an open-loop tour starting from current vehicle position and passing through viewpoints in $\mathbf Q$. We define a $n_q \times n_q$ cost matrix $\mathbb M_q$
describing the connection cost of each two elements in $\mathbb M_q$. The connection cost is evaluated using the length of a collision-free path searched by A* algorithm. Similar to FUEL, we formulate the problem as the Asymmetric Travelling Salesman Problem (ATSP), a variant of TSP, and solve it using the available algorithm~\cite{lin1973effective}. This section is referred to as the back-end of the exploration planner.

\subsection{Local Trajectory Generation}
\label{sec:local}
Given a collision-free global path generated in Sec.~\ref{sec:global}, we utilize a modified version of our previous work, BubblePlanner \cite{ren2022bubble}, to achieve smooth and energy-efficient local planning on LiDAR points. We represent the trajectory using a piecewise polynomial. The local planner first performs a batch sample algorithm along the path and generates a spherical safe flight corridor (SSFC). The corridor-constrained trajectory optimization problem is then solved using a spatial-temporal decomposition method \cite{wang2022geometrically}, which considers both smoothness and short trajectory execution time. The maximum velocity $v_\mathrm{max}$ and maximum acceleration $a_\mathrm{max}$ are also constrained to ensure kinodynamic feasibility. The local planner adopts a Receding Horizon Planning framework, with the planning horizon set to \SI{15}{m}. Replanning is triggered when the trajectory collides with newly sensed obstacles or when the global planner sends new targets.

\section{Experiments}

\subsection{Benchmark Comparison}
In this section, we present a comparative analysis of the proposed method and three state-of-the-art exploration algorithms, namely FUEL~\cite{zhou2021fuel}, GBP~\cite{dang2019graph}, and NBVP~\cite{bircher2016receding}. We assess the performance of these algorithms using a point-realistic simulator~\cite{kong2022marsim} in two large-scale scenarios, as depicted in Fig.~\ref{fig:map}. These scenarios were constructed by scanning real-world environments using LiDAR. We conducted four runs of all algorithms in each scenario, each with different initial configurations. It is important to note that all algorithms employed the same initial configurations in each run. We constrained the maximum speed of the UAV to $v_\mathrm{max} = 2.5m/s$ for all methods. The exploration process's time limit was set to 1200 seconds (20 minutes). All tests were conducted on an Intel Core i7-8700@3.2GHz CPU.

\begin{figure}[t]
 \centering 
 \includegraphics[width=0.47\textwidth]{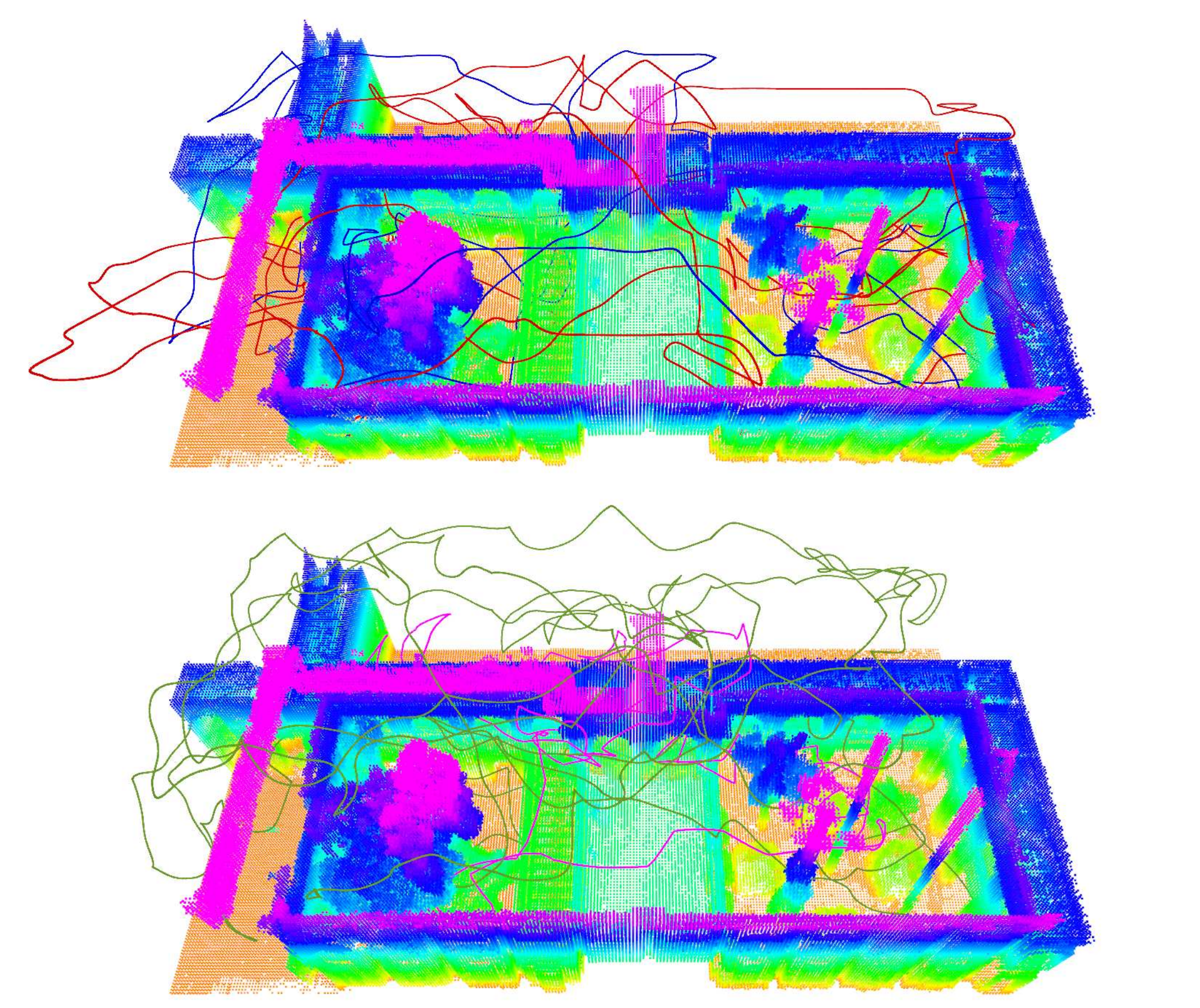}
 \caption{The executed trajectories of all benchmarked methods in \textit{Building} scenario. Top: The trajectory of the proposed method (blue) and FUEL~\cite{zhou2021fuel} (red). Bottom: The trajectory of GBP~\cite{dang2019graph} (green) and NBVP~\cite{bircher2016receding} (pink).}
 \label{fig:traj_main}
\end{figure}

\begin{table}[t]
 \small
 \centering
 \caption{Run Time Comparison}
 \setlength{\tabcolsep}{1mm}
 \begin{tabular}{@{}ccccccc@{}}\toprule
 \label{tab:run_time}

 Scene & \multicolumn{4}{c}{Methods average run time (s)}\\
 \cmidrule(r{4pt}){2-5}
 &Proposed & FUEL~\cite{zhou2021fuel} & GBP~\cite{dang2019graph} & NBVP~\cite{bircher2016receding}\\
 \toprule
 Building &\textbf{0.155} & 0.419 & 2.821 &7.456 \\
 Forest &\textbf{0.288} & 1.139 & 3.423 &10.078 \\
 Forest (MID360) &\textbf{0.313} & 1.467 & 5.438 &19.622 \\
 
\bottomrule \end{tabular}
 \end{table}

\cmmnt {\begin{table}[t]
 \small
 \centering
 \caption{Run Time Comparison}
 \setlength{\tabcolsep}{1mm}
 \begin{tabular}{@{}ccccccc@{}}\toprule
 \label{tab:run_time}

 \multirow{2}{*}{Test} &  & \multicolumn{4}{c}{Methods run time (s)}\\
 \cmidrule(r{4pt}){3-6}
 & &Proposed & FUEL~\cite{zhou2021fuel} & GBP~\cite{dang2019graph} & NBVP~\cite{bircher2016receding}\\
 \toprule
 \multirow{3}{*}{Test 1} 
 &Avg &\textbf{0.155} & 0.419 & 2.821 &7.456 \\
 &Max &1.363 & 6.757 & 4.594 &12.061 \\
 &Min &0.017 & 0.010 & 1.213 &2.996 \\ 
 \toprule 
 \multirow{3}{*}{Test 2} 
  &Avg &\textbf{0.289} & 1.139 & 3.423 &10.078 \\
 &Max &2.177 & 15.376 & 6.926 &15.127 \\
 &Min &0.059 & 0.032 & 1.891 &4.709 \\ 
 
\bottomrule \end{tabular}
 \end{table} }

\begin{figure}[t]
 \centering 
 \includegraphics[width=0.45\textwidth]{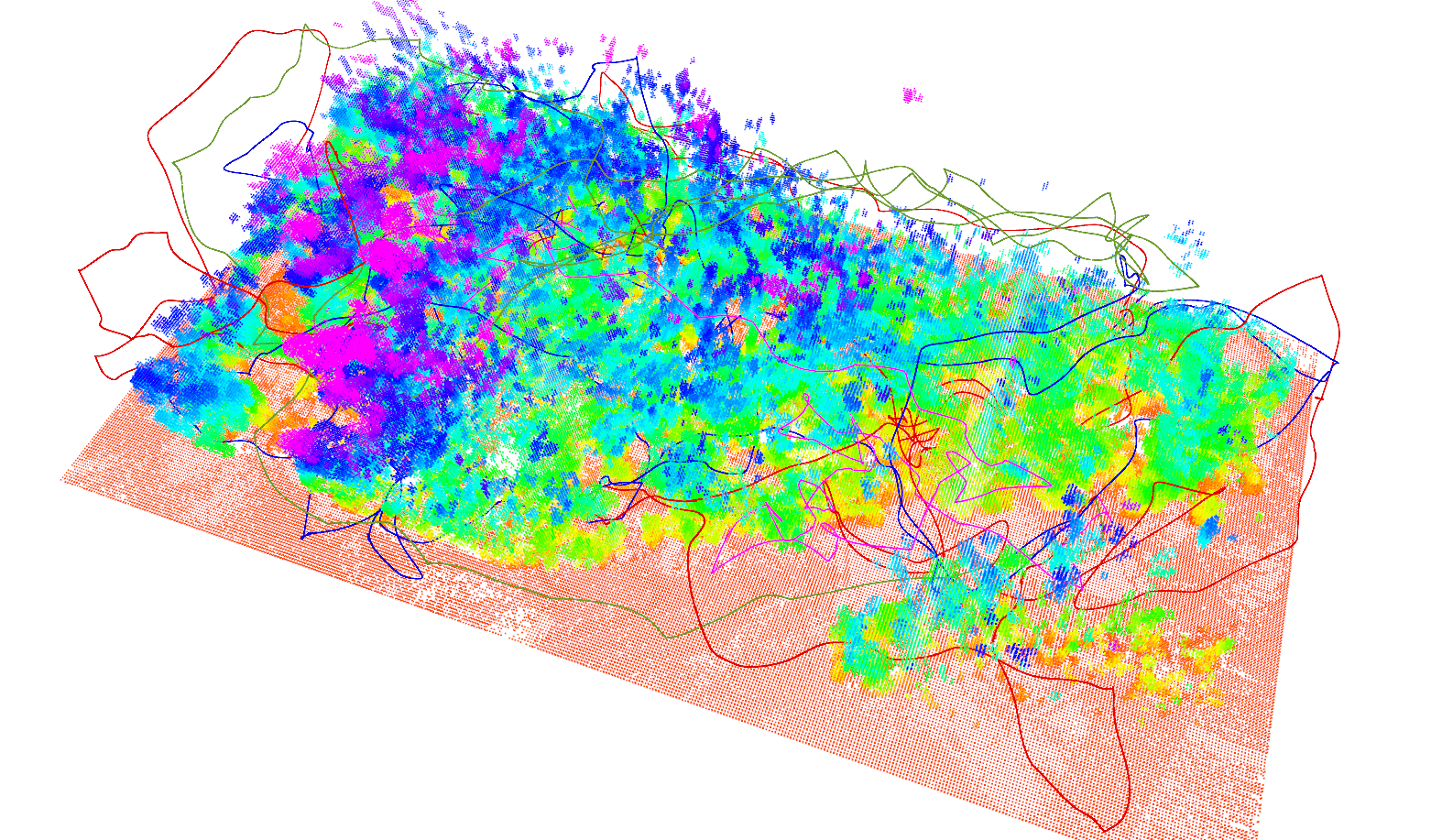}
 \caption{The executed trajectories of all methods in \textit{Forest} scenario: The proposed (blue), FUEL~\cite{zhou2021fuel} (red), GBP~\cite{dang2019graph} (green) and NBVP~\cite{bircher2016receding} (pink)}
 \label{fig:traj_forest}
\end{figure}

\begin{figure}[t]
 \centering 
 \includegraphics[width=0.47\textwidth]{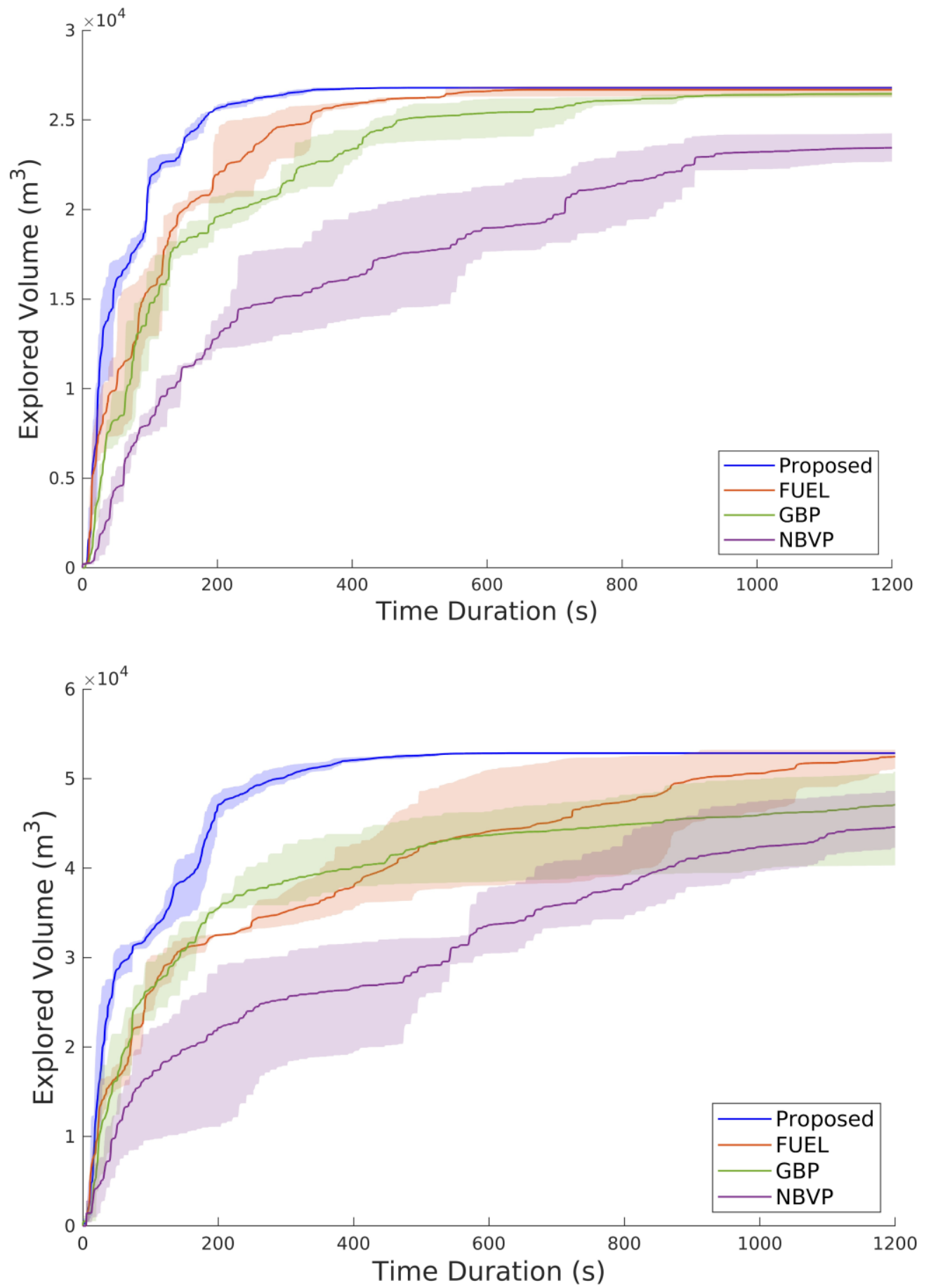}
 \caption{The explored volume over time in \textit{Building} scenario (top) and \textit{Forest} scenario (bottom)}
 \label{fig:main}
\end{figure}

\textit{1) Building:} We first conduct the exploration in a scenario that contains a two-story building surrounding a garden in the middle (Fig.~\ref{fig:map}(a)). Fig.~\ref{fig:main} shows the explored volume of all algorithms over time. The semi-transparent region in color is formed by the upper-bound and lower-bound of the four algorithm runs, while the solid line represents the mean of these four runs. The proposed method showcases higher exploration rate than all benchmarked methods throughout the entire exploration process. Fig.~\ref{fig:traj_main} displays the executed trajectory of all methods after exploration terminates. In this scenario, the proposed method achieves complete exploration with an average flight distance of 655.8m, while FUEL and GBP achieve 876.7m and 1060.3m respectively. 
NBVP is unable to achieve full exploration of the scene within the time limit. Table~\ref{tab:run_time} presents the overall computational time. The proposed method demonstrates performance 2.5+ times faster than FUEL, 18+ times faster than GBP, and 48+ times faster than NBVP.  

\textit{2) Forest:} The second test is conducted in a dense forest scenario (Fig.~\ref{fig:map}(b)). As shown in Fig.~\ref{fig:main}, all methods take longer to fully explore the environment. This is because the \textit{Forest} scenario is twice as large, more cluttered, and complex than \textit{Building}. Nevertheless, the proposed method demonstrates impressive performance. In this scenario, the proposed method completes exploration in an average of 579.8s with an average flight distance of 804.1m, FUEL completes in an average of 1139.6s and 1098.4m respectively. GBP and NBVP are unable to complete the exploration within the given time limit. 
In terms of computational time, our algorithm performs 4 times faster than FUEL, 11+ times faster than GBP, and 35 times faster than NBVP. 

Notably, GBP demonstrates a higher exploration rate than FUEL in the early stage of the exploration process by visiting high-gain unexplored regions in a greedy manner. However, it falls short in the late exploration stage due to the lack of global optimality. The proposed method combines greedy strategy and global optimization to explore a large volume in the early stage of the exploration process and then efficiently cover unexplored regions to achieve full exploration of the environment. 

\begin{figure}[t]
 \centering 
 \includegraphics[width=0.47\textwidth]{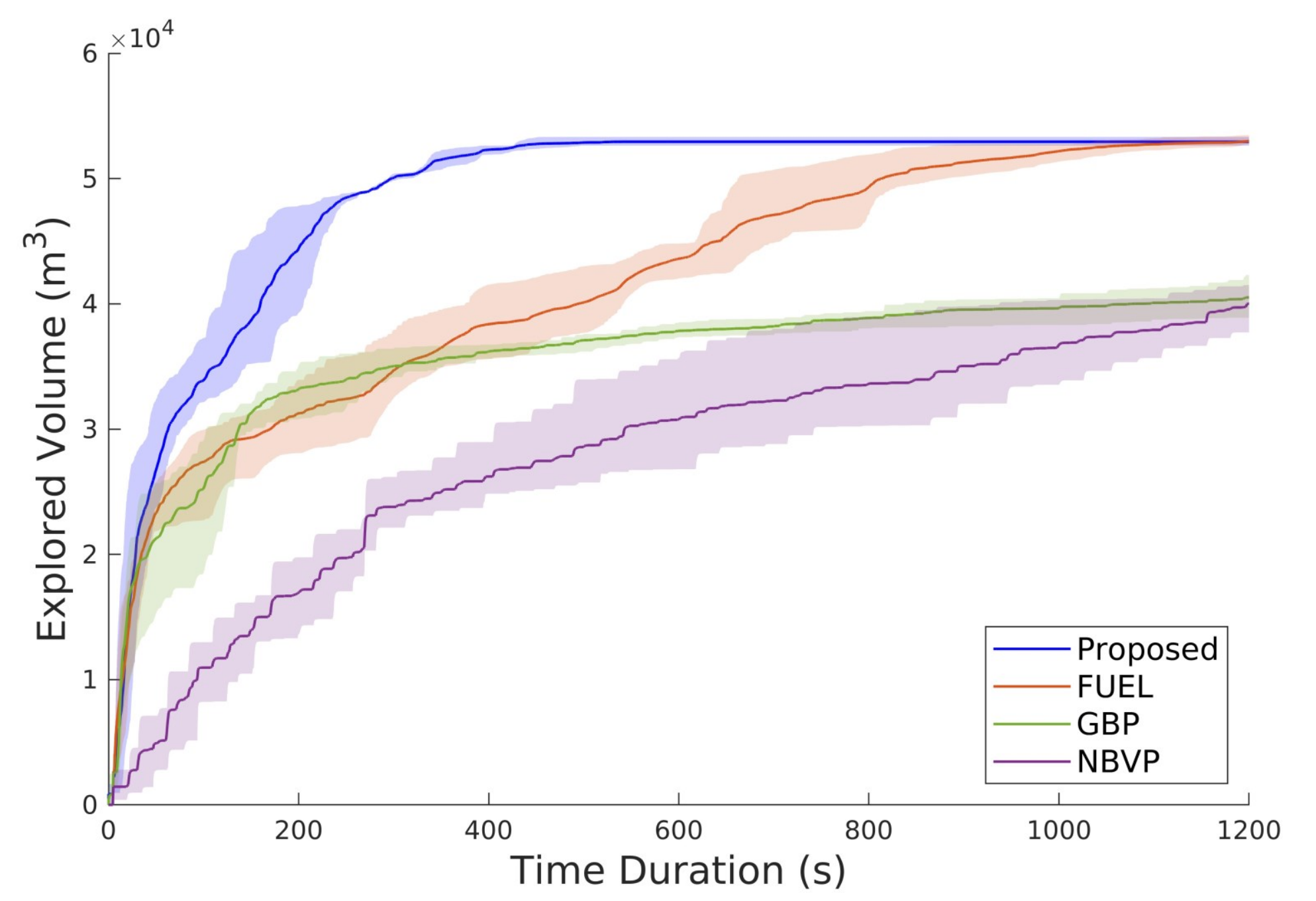}
 \caption{The explored volume over time in \textit{Forest (MID360)} scenario}
 \label{fig:forest_360}
\end{figure}

\begin{table}[t]
 \footnotesize
 \centering
 \caption{Computational Time of Components}
 \setlength{\tabcolsep}{1mm}
 \begin{tabular}{@{}cccccccc@{}}\toprule
 \label{tab:component_main}
\multirow{3}{*}{Scenes}  &\multirow{3}{*}{Methods}  & \multicolumn{3}{c}{Viewpoints generation} & \multicolumn{3}{c}{Global optimization}\\
 & & \multicolumn{3}{c}{(Front-end) (s)} & \multicolumn{3}{c}{(Back-end) (s)}\\
  \cmidrule(r{4pt}){3-5}\cmidrule{6-8} 
 & & Frontier & View. & Total & Cost. & TSP & Total \\
 \toprule
 \multirow{2}{*}{Building} & Proposed & 0.046 & 0.019 & 0.065 & 0.048 &0.009 & 0.057\\
 & FUEL~\cite{zhou2021fuel} & 0.069 & 0.078 & 0.147 & 0.210 &0.036 & 0.246\\
 \toprule
 \multirow{2}{*}{Forest} & Proposed &0.129 &0.036 & 0.165 & 0.046 &0.013 &0.059 \\
 & FUEL~\cite{zhou2021fuel} &0.208 &0.172 & 0.380 & 0.621 &0.115 &0.736 \\
 
 \bottomrule\end{tabular}
 \end{table}
 
 \cmmnt {\begin{table}
 \caption{Computational Time}
 \setlength{\tabcolsep}{1mm}
 \begin{tabular}{@{}ccccccc@{}}\toprule\midrule
 \label{tab:component_forest}
 \multirow{2}{*}{Methods} & \multicolumn{3}{c}{Viewpoints Generation} & \multicolumn{3}{c}{Global Optimization}\\
 & Frontier & Viewpoint & Total & Cost Mat. & TSP & Total \\
 \hline
 Proposed &0.129 &0.036 & 0.165 & 0.046 &0.013 &0.059 \\
 FUEL~\cite{zhou2021fuel}     &0.208 &0.172 & 0.380 & 0.621 &0.115 &0.736 \\
 \midrule\bottomrule\end{tabular}
 \end{table} }

To further benchmark our methods with FUEL in detail, we compare the computational time of each component of both methods. The statistics are presented in Table.~\ref{tab:component_main}. The results show that our method runs 2+ times faster on average than FUEL in the front-end and 5-10 times faster in the back-end.

\textit{3) Forest (MID360):} To demonstrate the versatility of our method across various types of LiDARs, we conducted simulation tests in the dense forest scenario using a 360-degree FoV LiDAR: Livox MID360. Results are presented in Fig.~\ref{fig:forest_360} and Table.~\ref{tab:run_time}, from which similar conclusions could be drawn.

\subsection{Real-world Experiments}

\begin{figure}[t]
 \centering 
 \includegraphics[width=0.47\textwidth]{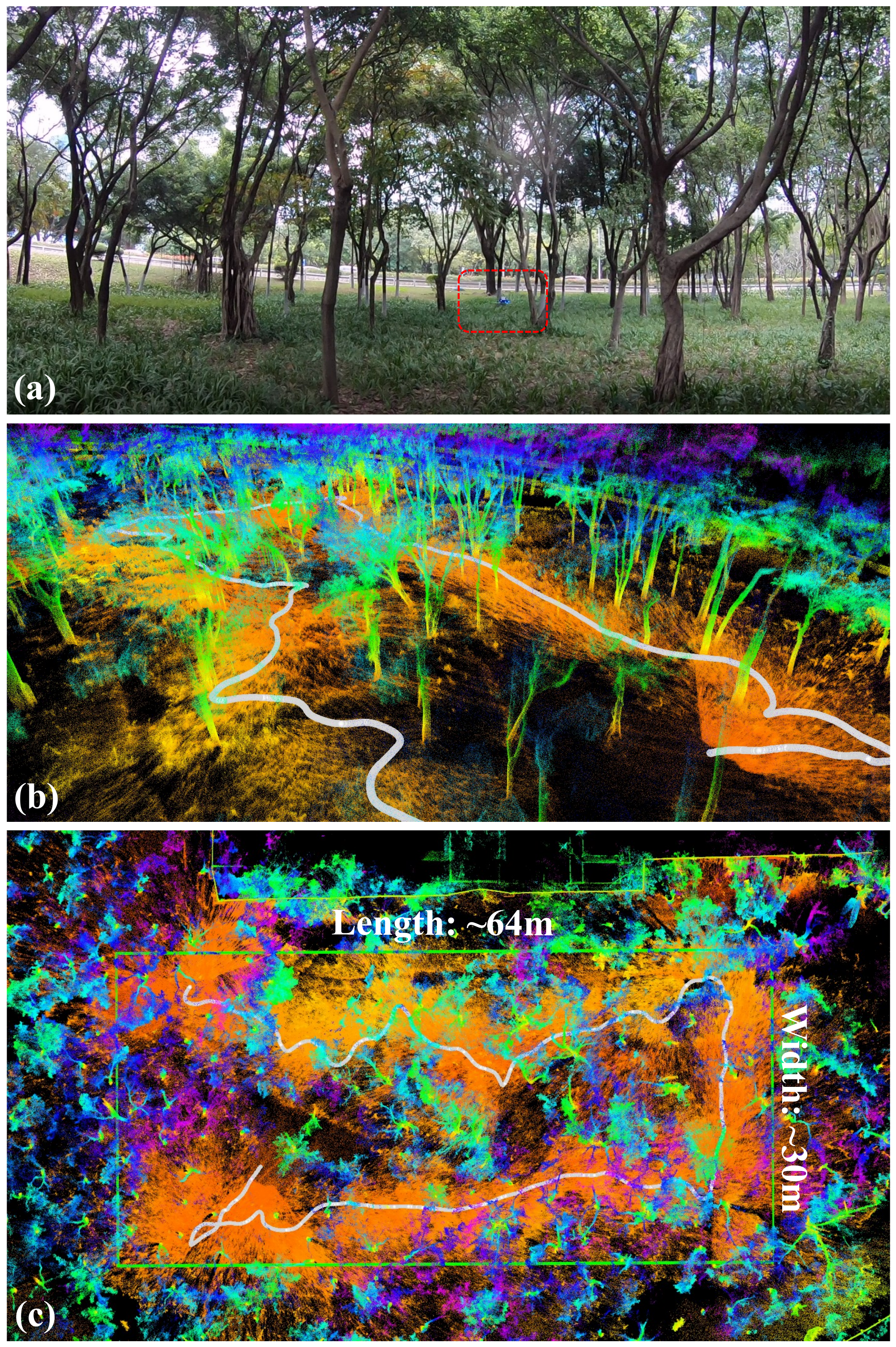}
 \caption{(a): Real-world experiment conducted in a forest. (b) and (c): Two different views of the online-built point cloud map and executed trajectory of the UAV. The green box is the bounding box of the area to be explored. Areas out of the bounding box were also observed due to the long LiDAR measuring range.}
 \label{fig:forest_real}
\end{figure}

Various real-world experiments are conducted to further validate our method. We build a LiDAR-based quadrotor platform. The platform is equipped with an Intel NUC onboard computer with CPU i7-10710U, Pixhawk flight controller, and LiDAR (Livox AVIA or Livox MID360). For localization and mapping, we rely on LiDAR and the built-in IMU of the flight controller to run~\cite{zhu2022decentralized}, a modified version of FAST-LIO2~\cite{xu2022fast}, providing high accuracy and high frequency state estimation. The time offset and extrinsic between LiDAR and IMU are calibrated by~\cite{zhu2022robust}. For trajectory following, we use an on-manifold model predictive controller~\cite{lu2022manifold}. 

\begin{figure}[t]
 \centering 
\includegraphics[width=0.47\textwidth]{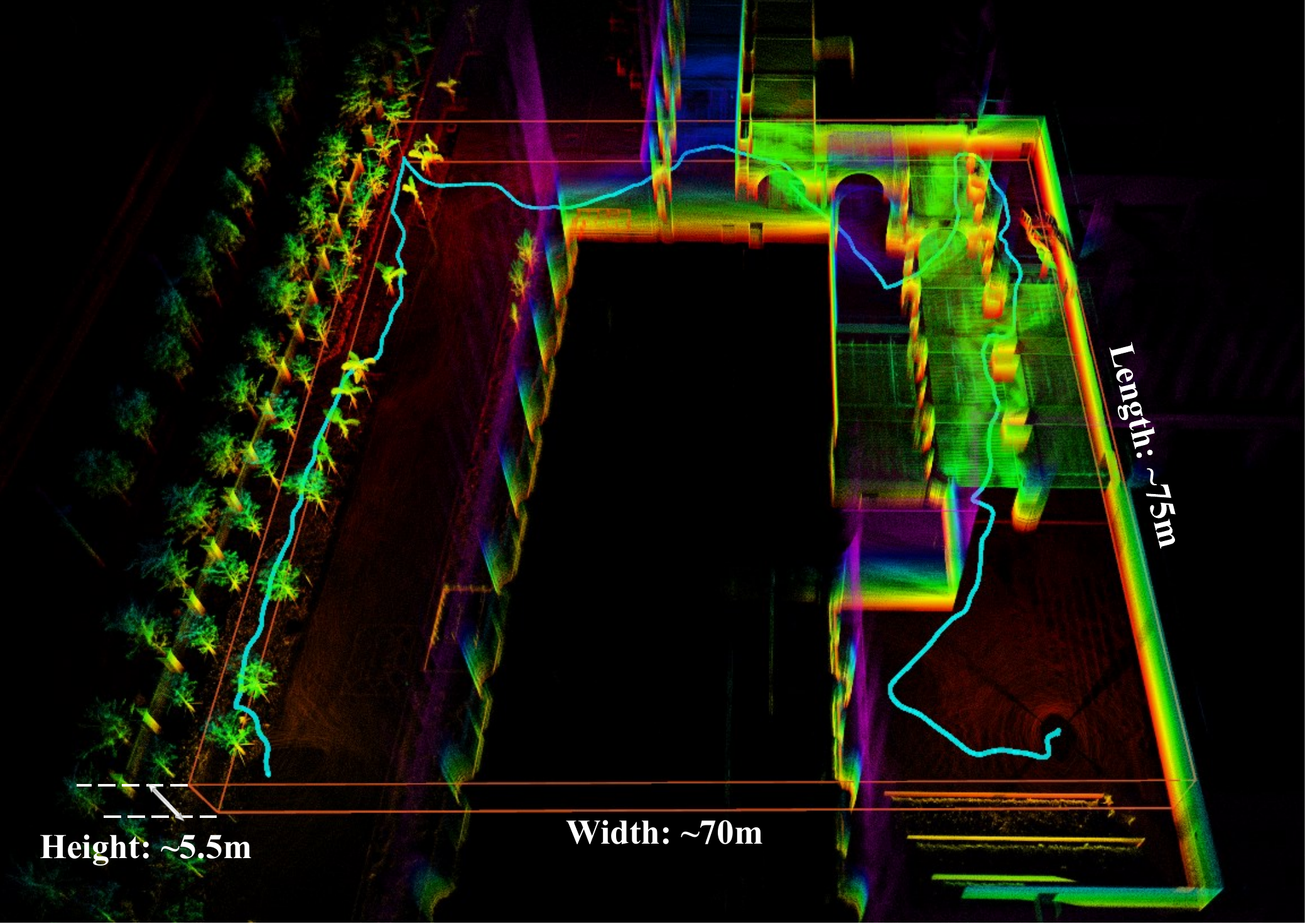}
 \caption{The overview of the online-built point cloud map and executed trajectory of the UAV. The orange box is the bounding box of the area to be explored. In the middle of the scene is a closed building, leading to no feasible path.}
 
 \label{fig:zhongda_over}
\end{figure}

\begin{figure}[t]
 \centering 
\includegraphics[width=0.47\textwidth]{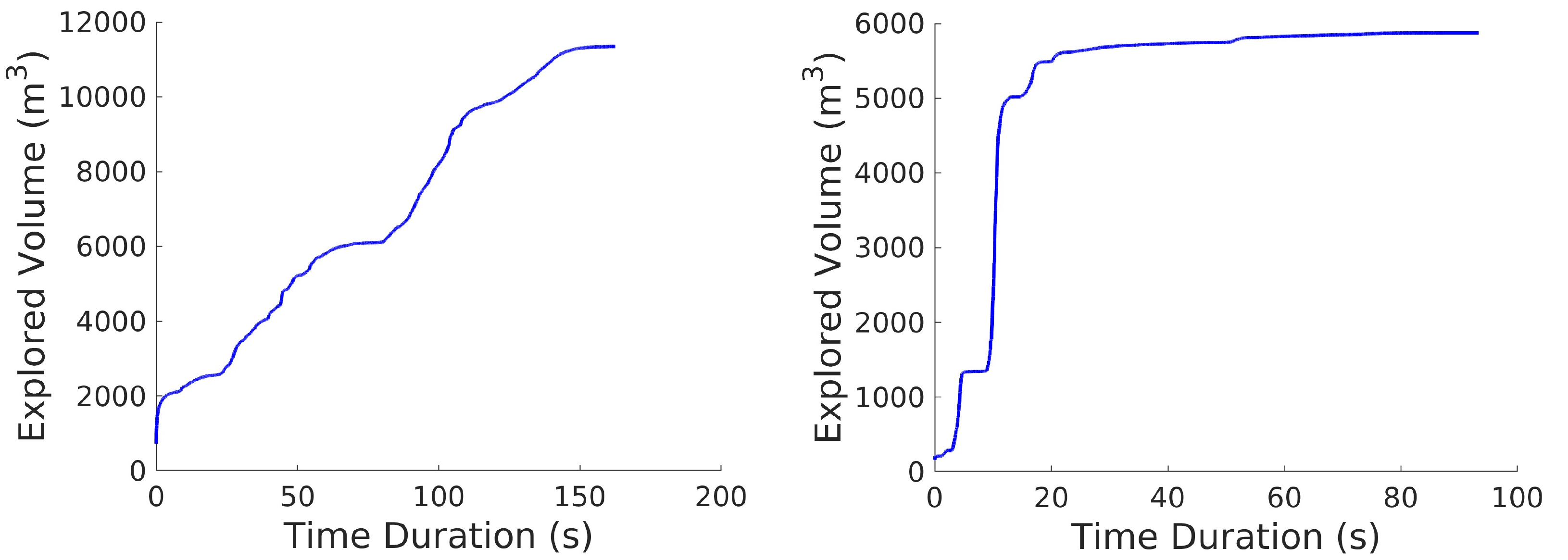}
 \caption{The explored volume over time in Scenario 1 (left) and Scenario 2 (right).}
 \label{fig:explore_real}
\end{figure}

First, we use the UAV to explore a large-scale and cluttered environment containing both indoor and outdoor spaces (Scenario 1). The size of the area to be explored is $[75 \times 70 \times 5.5]m^3$. 
In this scene, we equip the UAV with Livox MID360 LiDAR. In a representative run, the UAV completes the exploration in 155.8s, with a flight distance of 268.4m. The online-built point cloud map and the executed trajectory are displayed in Fig.~\ref{fig:cover} and Fig.~\ref{fig:zhongda_over}. The explored volume over time is shown in Fig.~\ref{fig:explore_real}. 
In this test, the UAV starts out at an open space. Then it explore in a hallway and proceed to outdoor space. 
The UAV had almost completed the exploration of the hallway after approximately 70 seconds, resulting in the flattening of the curve slope in Fig.~\ref{fig:explore_real}. At around 85 seconds, the UAV entered the outdoor space and covered a significant amount of unknown area, resulting in a steep increase in the curve slope. Finally, at approximately 160 seconds, the exploration of the outdoor space was completed, concluding the entire exploration process.
As observed in the results presented in Fig.~\ref{fig:zhongda_over}, the proposed method produced few revisit or back-and-forth planning motion throughout the entire exploration process.

Second, we use the UAV to explore a cluttered forest scene with a size of $[64 \times 30 \times 3.2]m^3$ (Scenario 2).
In this test the UAV is equipped with Livox AVIA LiDAR, which has a $[70.4^\circ \times 77.2^\circ]$ cone-shape FoV. The UAV takes 92s to complete the exploration with a flight distance of 159.2m. The point cloud and trajectory are displayed in Fig.~\ref{fig:forest_real} and statistics result is shown in Fig.~\ref{fig:explore_real}.

\section{Conclusion and Future Work}
In this paper, we proposed a novel method to support efficient autonomous exploration in large-scale and cluttered 3-D environments. In the front-end, we introduced the novel concept of an occlusion-free sphere to generate high-quality viewpoints at low computational cost. In the back-end, our method adopts a novel strategy that combines greedy with global optimization. The proposed method demonstrated a significant improvement in exploration efficiency and computational time savings.
Extensive simulation and real-world experiments showcased the outstanding performance of our method in large-scale, cluttered, and complex environments. In the future, the method can be extended to support the exploration of a larger scene by implementing it on multiple UAVs.

\section*{Acknowledgment}
This work is supported by the Grants Committee Early
Career Scheme of The University of Hong Kong under
Project 27202219, General Research Fund of Hong Kong
under project 17206920, and a DJI research donation. The authors gratefully acknowledge DJI for funding and Livox Technology for equipment support. The authors would like to thank Dr. Ximin Lyu of Sun Yat-sen University for the experimental site support during the whole work.

{\small
\bibliographystyle{unsrt}
\bibliography{reference}
}

\end{document}